\documentclass{article}

\usepackage{microtype}
\usepackage{graphicx}
\usepackage{subfigure}
\usepackage{booktabs} 
\usepackage{amsmath}
\usepackage{enumitem}
\usepackage{subcaption}

\setlist{nolistsep}

\usepackage{hyperref}

\usepackage[accepted]{mlsys2025}

\mlsystitlerunning{Checkpoint Sparsification and Quantization}

\begin{document}

\twocolumn[
\mlsystitle{BitSnap: Checkpoint Sparsification and Quantization in LLM Training}



\mlsyssetsymbol{equal}{*}

\begin{mlsysauthorlist}
\mlsysauthor{Yanxin Peng}{equal,infini}
\mlsysauthor{Qingping Li}{equal,infini}
\mlsysauthor{Baodong Wu}{infini}
\mlsysauthor{Shigang Li}{bupt}
\mlsysauthor{Guohao Dai}{sjtu}
\mlsysauthor{Shengen Yan}{tu}
\mlsysauthor{Yu Wang}{tu}
\end{mlsysauthorlist}

\mlsysaffiliation{infini}{InfiniGence, Beijing, Beijing, China}
\mlsysaffiliation{bupt}{Beijing University of Posts and Telecommunications, Beijing, Beijing, China}
\mlsysaffiliation{sjtu}{Qing Yuan Research Institute, Shanghai Jiao Tong University, Shanghai, China}
\mlsysaffiliation{tu}{Department of Electronic Engineering, Tsinghua University, Beijing, China
}

\mlsyscorrespondingauthor{Yanxin Peng}{peterpen@usc.edu}
\mlsyscorrespondingauthor{Qingping Li}{qingpingli@smail.nju.edu.cn}
\mlsyscorrespondingauthor{Baodong Wu}{wubd.cs@gmail.com}

\mlsyskeywords{Machine Learning, Large Language Model, Checkpoint Quantization, Checkpoint Sparsification}

\vskip 0.3in

\begin{abstract}

As large language models (LLMs) continue to grow in size and complexity, efficient checkpoint saving\&loading has become crucial for managing storage, memory usage, and fault tolerance in LLM training. The current works do not comprehensively take into account the optimization of these several aspects. This paper proposes a novel checkpoint sparsification and quantization method that adapts dynamically to different training stages and model architectures. We present a comprehensive analysis of existing lossy and lossless compression techniques, identify current limitations, and introduce our adaptive approach that balances compression ratio, speed, and precision impact throughout the training process. 
Experiments on different sizes of LLMs demonstrate that our bitmask-based sparsification method achieves 16x compression ratio without compromising model accuracy. Additionally, the cluster-based quantization method achieves 2x compression ratio with little precision loss.

\end{abstract}
]



\printAffiliationsAndNotice{\mlsysEqualContribution} 

\section{Introduction}

Large Language Models (LLMs) have grown exponentially in size and complexity in recent years, with state-of-the-art models now containing billions of parameters, such as GPT-3 (175B) \cite{brown2020language}, PaLM (540B) \cite{chowdhery2022palm}, LLaMA 3.1 (405B) \cite{vavekanand2024llama}, and Step-2 (1T). This rapid increase in model size has introduced significant challenges in model training, particularly during the checkpointing process, which becomes increasingly time-consuming. For example, training OPT (175B) \cite{zhang2022opt} involved processing 300B tokens over 57 days using 992 Nvidia A100 GPUs, while training LLaMA 3.1 (405B) required 54 days to process 15T tokens on 16K Nvidia H100 GPUs. Even with comprehensive fault-tolerant mechanisms, the training process for such large-scale models is still time-intensive. This leads to two major storage-related challenges: insufficient GPU memory and the excessive size of checkpoints, which creates I/O bottlenecks.

Previous works, such as those focusing on activation recomputation and flash attention, have primarily addressed the reduction of GPU memory footprint. However, our paper targets another critical issue: reducing checkpoint size. We have calculated and found that checkpoint size scales proportionally with model size in mixed precision training \cite{samsi2020compute}. Model states are stored using fp16 format while optimizer states are stored in fp32 format. For instance, the checkpoint size for GPT-3 (175B) amounts to 2.3TB, and the theoretical write speed of NVMe M.2 SSD is up to 3500 MB/s, and assuming no bandwidth competition, it would still take approximately 11 minutes to complete the storage process. Table \ref{table:llm_ckpt_time} shows a list of public models checkpoint save time. To optimize effective training time, researchers often reduce checkpoint saving frequency (typically saving every few hundred iterations). However, in large-scale training environments with 1,000+ or even 10,000+ GPUs, the frequency of hardware failures is high (e.g., OPT training experienced 2 machine crashes per day, and LLaMA 3.1 training saw 8 crashes per day). Infrequent checkpointing in such environments can lead to significant losses in training progress when faults occur. Therefore, there is an urgent need for a rapid checkpoint-saving strategy to mitigate these challenges. 

\begin{table*}[h]
\centering
\caption{Summary of Popular Large Language Models}
\label{table:llm_ckpt_time}
\begin{tabular}{lccc}
\toprule
Model & Number of Parameters & Approximate Time to Save Checkpoint & Year Published \\
\midrule
PaLM 540B              & 540 billion           & ~34.5 minutes                        & 2022           \\
Llama3.1 405B               & 405 billion           & ~25.1 minutes                        & 2024           \\
GPT-3 175B              & 175 billion           & ~10.8 minutes                        & 2020           \\
OPT 175B            & 175 billion           & ~10.8 minutes                       & 2023           \\
LLaMA-2 70B         & 70 billion            & ~4.3 minutes                         & 2023           \\
LLaMA-2 13B         & 13 billion            & ~0.8 minutes                         & 2023           \\
GPT-2 XL               & 1.5 billion           & ~0.1 minutes                         & 2019           \\

\bottomrule
\end{tabular}
\end{table*}


To address the growing challenge of increasing checkpoint size in large-scale model training, a variety of innovative approaches have been proposed, including Transom \cite{wu_transom_2023}, DLRover \cite{wang2024dlrover}, Megatron-LM \cite{shoeybi_megatron-lm_2020}, Gemini \cite{wang2023gemini}, and others. These methods commonly employ asynchronous checkpointing mechanisms designed to minimize the impact of checkpointing on overall training time. Specifically, these approaches utilize a daemon process that operates concurrently with the main training process, managing the task of persisting checkpoints to storage without interrupting the training workflow. While this strategy effectively reduces the time perceived by users for checkpointing, it also results in increased memory consumption. This trade-off highlights the need for more advanced and efficient compression techniques to manage memory usage without compromising the benefits of reduced checkpointing time.

Therefore, in order to propose an effective checkpointing strategy, we are going to focus on discussing the optimal compression strategy for different stages of pre-training a model in order to have the best efficiency and the most robust fault tolerance performance. 

We develop a checkpoint engine named BitSnap to expedite the checkpoint save and load time during LLM training. BitSnap employs the checkpoint sparsification and quantization method to handle model states and optimizer states within a single checkpoint, resulting in a memory/storage compression ratio of 16x and 2x respectively. Subsequently, we utilize in-memory redundancy to enable the LLM model to load checkpoints from in-memory checkpoints in most instances instead of loading from low-speed disks. In summary, we have made the following contributions.

\begin{itemize}
    \item We build a checkpoint engine to enable asynchronous checkpoint saving, which decrease the saving time from minutes or hours to seconds. And we use in-memory redundancy to enable LLMs' recovering loads checkpoint from memory as far as possible. In-memory redundancy will save a number of iterations in memory and training will be recovered from the latest available checkpoint iteration. Combined with our sparsification methods, it will not occupy too many memories.
    
    \item We design a bitmask-based sparsification method for model states, we firstly save a base checkpoint, and for the next numbers of checkpoints we only save the delta value on top of the base checkpoint. Furthermore, we use bitmask method to compress the delta value, which has a great save on model states size.
    
    \item We design a cluster-based quantization method for optimizer states. Tensor values into will firstly be splited into many clusters, and then for each cluster, each value will be represented by the most similar value in uint8 value range. These two compression methods lead to a great storage degradation. We employs the compression methods on our checkpoint engine, and the result on GPT models shows our compression methods contributes to 2x compression ratio on optimizer states and 16x compression ratio on model states.
\end{itemize}


\section{Background and Related Works}

\subsection{Asynchronous Checkpoint}
As LLMs grow larger and larger, checkpoint is becoming too large to save in a short time. Transom \cite{wu_transom_2023}, DLRover \cite{wang2024dlrover}, Megatron-LM \cite{shoeybi_megatron-lm_2020}, Gemini \cite{wang2023gemini}, MegaScale \cite{jiang2024megascale} are employed asynchronous checkpoint in their LLM training. They leverages the fact that memory bandwidth is higher than storage bandwidth and different nodes will not compete for memory bandwidth. In this method, checkpoints are initially copied from GPU memory to CPU memory(device to host, D2H) before being gradually persisted to storage through an asynchronous process. In asynchronous checkpoint, a few seconds will still be used in D2H. Gemini applies more strategies in asynchronous saving, it overlaps the D2H checkpoint copy and the training process, which totally destroy the checkpoint time during training process.

\subsection{Model Compression}

During the pre-training phase, selecting appropriate checkpointing strategies is crucial for balancing loss and gain. An effective checkpointing strategy must carefully manage the trade-off between checkpoint size and precision loss, as these two factors are often in conflict. The size of the checkpoint impacts I/O speed, while precision loss affects the model's ability to recover from failures. Therefore, it is essential to apply different compression techniques at various stages of pre-training, as the model's characteristics evolve throughout the process.

Checkpoint compression plays a key role in this strategy by aiming to reduce the storage footprint and I/O overhead of saving checkpoint, without significantly affecting model performance or training dynamics. To achieve this, current research has developed a range of compression methods that can be broadly categorized into two types: lossy and lossless, each offering distinct trade-offs and applications. By strategically integrating both lossy and lossless techniques at different stages of the compression pipeline, state-of-the-art (SOTA) approaches seek to optimize overall performance while adapting to the changing demands of the pre-training process.

\begin{figure}[h]
  \centering
  \includegraphics[height=200pt]{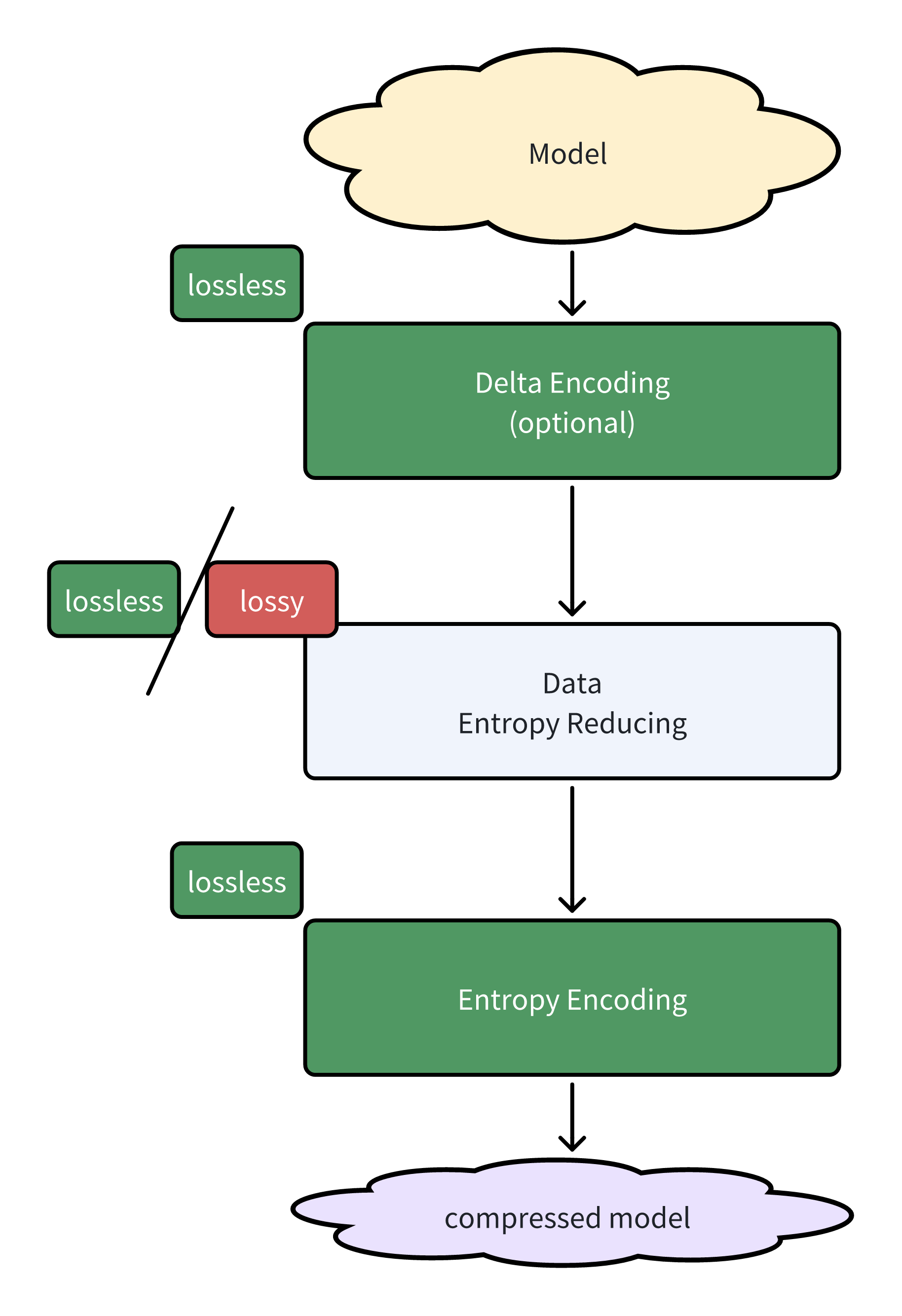}
  \caption{Compression Pipeline.}
  \label{fig1_compression_pipeline}
\end{figure}

As illustrated in Fig.~\ref{fig1_compression_pipeline}, the SOTA checkpoint compression pipeline typically involves several stages, including delta encoding, entropy reduction, and entropy encoding. The fundamental strategy, as synthesized from various approaches \cite{chen_efficient_2020, li_excp_2024, li_memory_2023, shi_lossy_2023, agrawal_dynaquant_2023, shen2024efficient, ibrahim2024jit}, is to apply techniques that reduce the entropy of the data before employing entropy encoding methods, such as Huffman Encoding, to further compress the data.

\begin{figure}[h]
  \centering
  \includegraphics[width=0.5\textwidth]{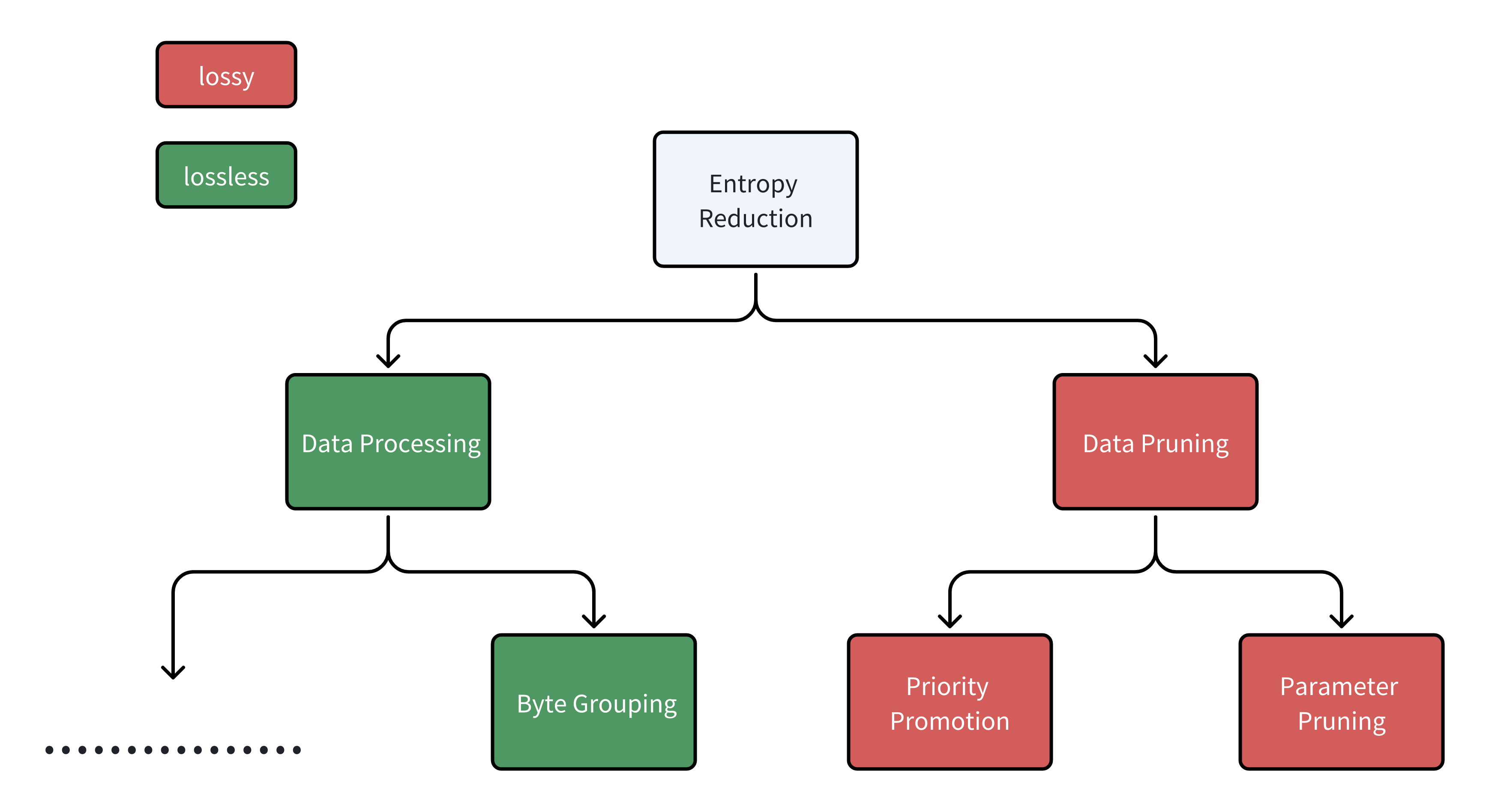}
  \caption{Entropy Reduction Approaches.}
  \label{fig2_entropy_reduction}
\end{figure}

These methods uniformly employ entropy encoding as the final step in the compression process, following a preliminary entropy reduction phase. The key difference across approaches lies in how aggressively they reduce entropy. For instance, some methods adopt more aggressive entropy reduction strategies \cite{li_excp_2024}, while others prioritize conservative or lossless entropy reduction techniques \cite{hershcovitch_lossless_2024}, as depicted in Fig.~\ref{fig2_entropy_reduction}. The recently published work AWQ \cite{lin2024awq} discovers the model weights that are particularly crucial for model performance and refrains from quantizing these model weights to mitigate precision loss.

\subsubsection{Lossy Compression Techniques}
Lossy compression methods prioritize high compression ratios at the cost of some precision loss. These techniques often employ strategies such as pruning, quantization, and delta encoding to achieve significant size reductions:

\begin{itemize}
    \item \textbf{Pruning}: This involves removing less important neurons or weights from the model. Methods like those proposed in \cite{ma_llm-pruner_2023, chen2020efficient, chekuri2012pruning, zhu2017prune} demonstrate how structured and unstructured pruning can reduce model size with minimal impact on performance.
    
    \item \textbf{Quantization}: By reducing the precision of model parameters from high-precision (e.g., 32-bit floating-point) to lower-precision formats (e.g., 8-bit integers), quantization techniques like those in \cite{dettmers_8-bit_2022, polino2018model, zhou2018adaptive} achieve substantial compression ratios.            
    
\end{itemize}

While these lossy methods can achieve impressive compression ratios—sometimes up to 70x as reported by ExCP \cite{li2024excp}, they will encounter the situation of a sudden jump of loss if the continued training from a breakpoint, costing hundreds or thousands of iterations to recover. This will ruin the training results gained from previous epoches and will potentially causing the explosion of gradient. Therefore, this aggressive way of parameter pruning will be in the tradeoff of fault tolerance, and they often require careful tuning to balance compression with model accuracy and training stability. 

\subsubsection{Lossless Compression Techniques}
Lossless compression ensures that the original model state can be fully reconstructed without any loss of information. Recent work by MIT researchers \cite{hershcovitch_lossless_2024} has shown that even without introducing any loss, foundation models like GPT-2 can achieve compression rates of up to 21.9\% through input data processing techniques like byte grouping, which utilizes the characteristics of the exponent part to group parameters to achieve entropy coding.

These techniques often involve:
\begin{itemize}
    \item \textbf{Delta Encoding}: Approaches such as LC Checkpoint \cite{chen_efficient_2020} and DynaQuant \cite{agrawal_dynaquant_2023} leverage the concept of incremental storage, focusing on compressing the changes between checkpoints rather than the entire model state. 
    
    \item \textbf{Entropy Coding}: Methods like Huffman encoding are applied after preprocessing the data to reduce entropy.
    
    \item \textbf{Byte-level Analysis}: By distinguishing between different parts of floating-point representations (sign, exponent, mantissa), more efficient compression schemes can be devised.
\end{itemize}
\subsubsection{Optimizer State Compression}


The optimizer states constitutes a significant portion of the checkpoint size. To reduce the storage footprint, it is essential to compress these optimizer states. Recent advancements, including the 8-bit dynamic quantization proposed by Dettmers \cite{dettmers_8-bit_2022}, \cite{dettmers_8-bit_2016}, the approximate K-means method by Agrawal \cite{agrawal_dynaquant_2023}, and the extreme pruning approach by Li \cite{li_excp_2024}, have done much work on optimizer states.

Dettmers' 8-bit dynamic quantization is an effective technique for reducing both the memory footprint on GPUs and the storage footprint of checkpoints, achieving nearly a 75\% reduction in checkpoint size. Agrawal's method uses an approximate K-means clustering to dynamically quantize blocks, while Li's approach aggressively prunes trivial optimizer states. However, Agrawal's K-means-based method, despite reducing runtime through approximation, still requires considerable time to cluster optimizer states. On the other hand, Li's aggressive pruning method can lead to abnormal loss behavior when resuming from a checkpoint, necessitating additional computational resources for recovery.

\section{BitSnap Design}
\begin{figure*}
    \centering
    \includegraphics[height=300pt]{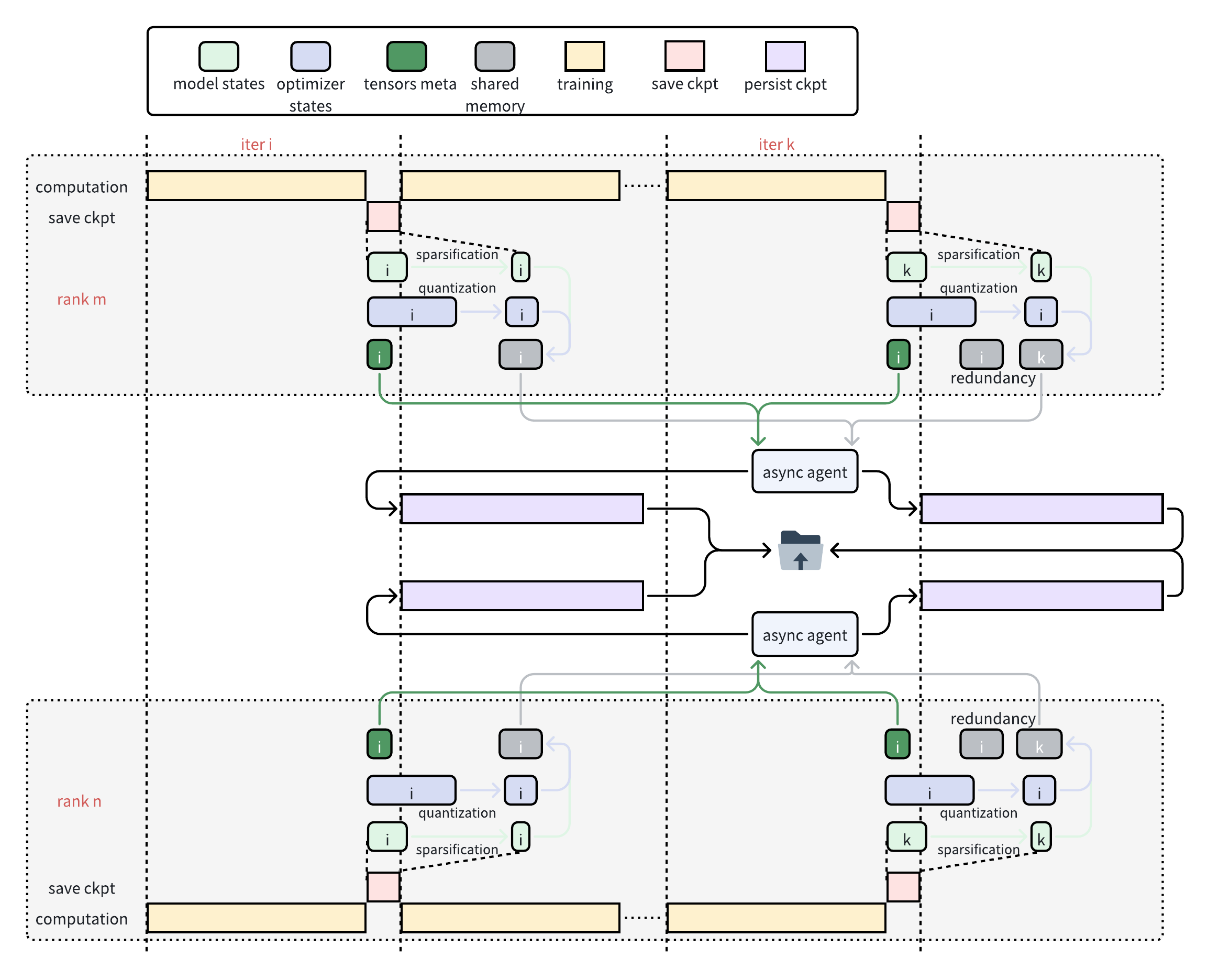}
    \caption{Our checkpoint engine structure}
    \label{fig:ckpt_structure}
\end{figure*}

\subsection{Overview}

In our asynchronous checkpoint engine (see Fig. \ref{fig:ckpt_structure}), the async agent operates as a daemon process communicating with the main training process. In addition to the async agent, another key aspect is the transfer of model data from GPU memory to CPU memory. This step is vital for ensuring that checkpoint data is properly preserved and available for future use. 

Firstly, model data including model states and optimizer states of a particular checkpoint step will be sparsified and quantized into a non-memory-consuming data. Then the compressed data will be copied to shared memory, and model metadata will be sent to async agent. Then the checkpoint step in training process is done and the next iteration can be launched immediately. Asynchronously, async agent will read the metadata and persist the compressed data into storage. Besides, our memory redundancy method will have multiple checkpoint iterations in shared memory, in case of the latest checkpoint iteration is broken we can recover from the former checkpoint iteration.

We will introduce our asynchronous checkpoint engine in \ref{sec:async-ckpt-engine}. Our proposed compression method enhances existing techniques by integrating novel approaches to balance compression efficiency with model accuracy throughout the training process. For lossless compression, we draw inspiration from sparse matrix compression methods and optimize them into a bitmask-based sparsification technique. This method proves especially effective for compressing model states in pre-training and post-training stage. For lossy compression, we use cluster-based quantization method to compress tensors into uint8 data type. This is useful for compress optimizer states due to their high precision. We will explain the lossless and lossy compression in Sec. \ref{sec:bitmask-compression} and Sec. \ref{sec:quantization-compression} respectively.

\subsection{Asynchronous Checkpoint Engine}
\label{sec:async-ckpt-engine}

We construct an asynchronous checkpoint engine as illustrated in Fig. \ref{fig:ckpt_structure}. Firstly, we write the model checkpoint into shared memory. Subsequently, the training process will continue to run. The async agent will manage to persist the checkpoint in shared memory to storage. Owing to the high memory bandwidth and multi-node parallel writing, saving a checkpoint into memory incurs less cost than saving it to storage.


However, in the event that the checkpoint of the current iteration is either not completely written to shared memory or is invalid, during recovery, loading can only be performed from disk, thereby resulting in a longer recovery time. To address this issue, an in-memory redundancy approach has been designed. This approach saves checkpoints of multiple iterations in memory. When the training process crashes and restarts and reads the checkpoint from the async agent, if the latest checkpoint is found to be damaged, recovery will be performed from the previous checkpoint saved in shared memory.

\begin{figure}[h]
  \centering
  \includegraphics[width=0.5\textwidth,height=0.5\textwidth]{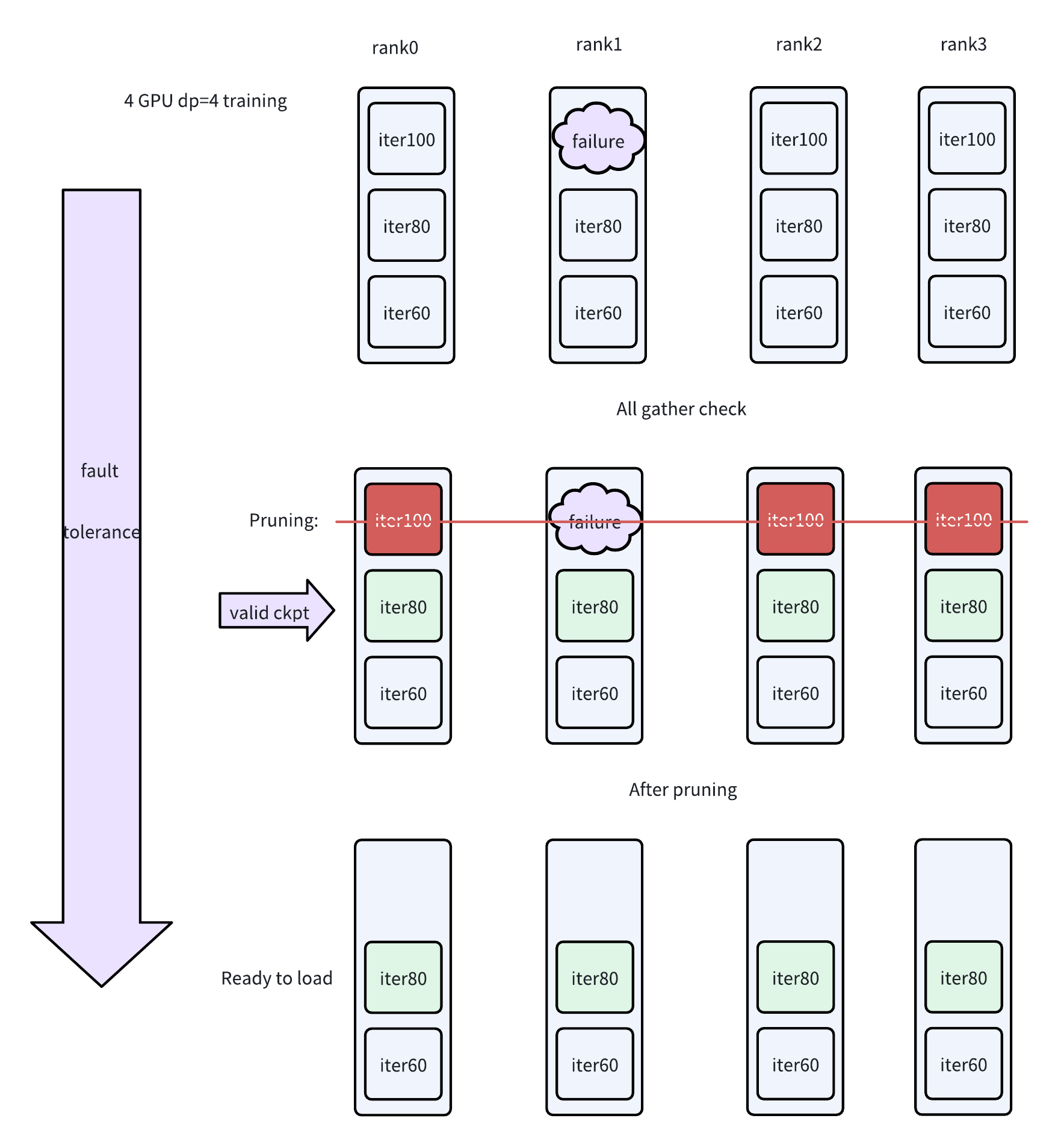}
  \caption{In-memory Redundancy Procedure}
  \label{fig:memory-redundancy}
\end{figure}

The entire process is depicted in Fig. \ref{fig:memory-redundancy}. There are four GPU ranks in training, and the checkpoint save interval is set to 20. The training process is currently in iteration 100, and checkpoints at iterations 80 and 60 have been copied into shared memory. However, rank 1 fails to copy its model data at iteration 100 into shared memory, resulting in the restart of the entire training. Subsequently, the training processes load the checkpoint from the async agent. Firstly, an all-gather check is performed to determine that the latest checkpoint iteration among all ranks is 100, except for rank 1 which returns 80. Then, the broken checkpoint at iteration 100 is pruned and loading is performed from iteration 80.

Combining in-memory redundancy with checkpoint compression can attain a reduced memory footprint and avert the issue of insufficient memory.

\subsection{Bitmask Based Sparsification for Model States} 
\label{sec:bitmask-compression}

We have observed that in mix precision training (model states are stored in fp16/bf16 types, while optimizer states are stored in fp32 types), when the loss remains stable, there is minimal change in the value of model states. For instance, the difference between iteration 500 and iteration 501 of GPT-2 Medium is only 15\%. This provides us with a significant opportunity for compressing model states. Therefore, firstly, we compute the difference between two checkpoint iterations to obtain the delta states. Based on these delta states, we explore several methods for compressing model states.





\textbf{Naive Bitmask Sparsification:} Sparse matrix additionally store the rows and columns of the non-zero values. In fact, we can reduce space of the additional elements by involving a bitmask. Firstly, we flatten the delta states matrix and build an uint8 bitmask list with the same shape of the flattened delta states. Then each element is valued 0 or 1 when the element of the same index in the flattened delta states is zero or non-zero. We will save the bitmask list and the nor-zero elements. The space required is $n+2n_c$. 

\begin{equation}
\label{eq:bitmask-required-space}
\begin{split}
    n+2n_c & < 2n   \\
    n_c & < \frac{1}{2} n   \\
\end{split}
\end{equation}

Bitmask compression is beneficial when the percentage of changed parameters is less than 50\% of the total number of parameters as shown in Eq. \ref{eq:bitmask-required-space}.

\textbf{Improved Bitmask Sparsification:} The naive bitmask sparsification method can be further enhanced by packing the bitmask into a bit sequence, reducing its size to $\frac{1}{8}$ of the original bitmask. The total space occupied by this improved bitmask is given by $\frac{1}{8} n+2n_c$.

\begin{equation}
\label{eq:improved-bitmask}
\begin{split}
    \frac{1}{8} n+2n_c & < 2n   \\
    n_c & < \frac{15}{16} n   \\
\end{split}
\end{equation}

\begin{figure}[h]
  \centering
  \includegraphics[height=150pt]{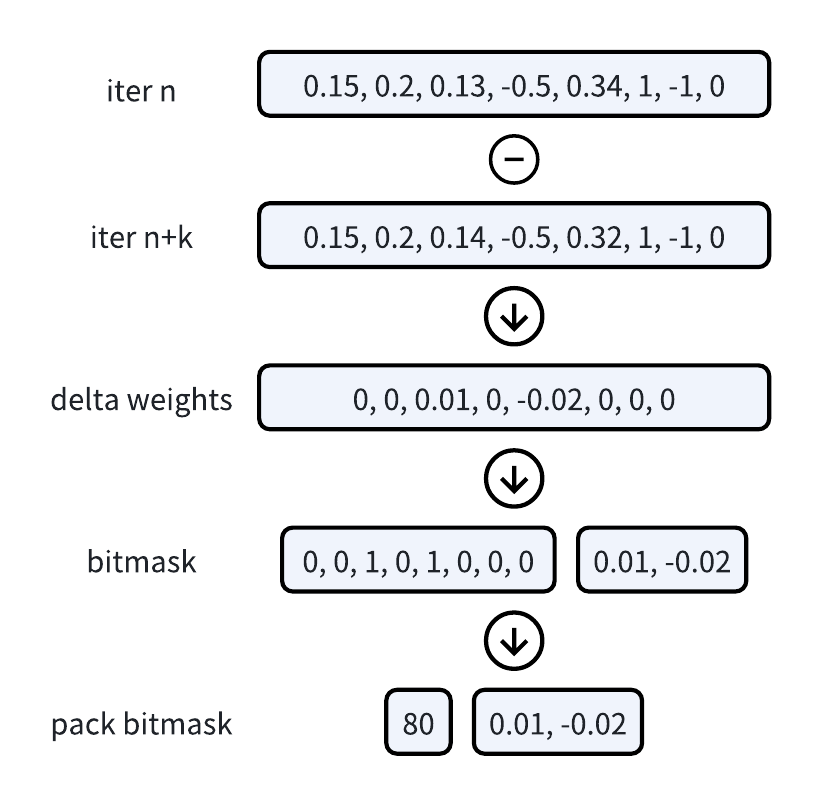}
  \caption{Improved bitmask sparsification}
  \label{fig:bitmask-compression-method}
\end{figure}

As expressed in Eq. \ref{eq:improved-bitmask}, this result implies that our improved bitmask sparsification method is beneficial if the changed portion of the parameters is smaller than 93.75\% of the total number of parameters. Our improved bitmask sparsification is shown in Fig. \ref{fig:bitmask-compression-method}. For instance, when the delta change amounts to 15\%, we can attain nearly a 5x lossless compression ratio.

\textbf{Rationale for Not Using Huffman Encoding:}
We decided against using Huffman encoding for two main reasons:
\begin{itemize}
    \item We did not pre-process the data to reduce its entropy. As suggested in \cite{hershcovitch2024lossless}, byte grouping could be applied to further reduce the size of the incremental checkpoint, but this would increase time consumption.
    \item Huffman encoding's efficiency would not surpass that of a packed bit sequence. This is because Huffman encoding typically represents only the most frequent symbol with a one-bit code, while the remaining symbols require at least two bits per element. Therefore, using Huffman encoding will not result in furthere gain of compression ratio.
\end{itemize}

\subsection{Cluster Based Quantization for Optimizer States} 
\label{sec:quantization-compression}

\begin{figure*}[h]
  \centering
  \includegraphics[width=1\textwidth,height=0.3\textwidth]{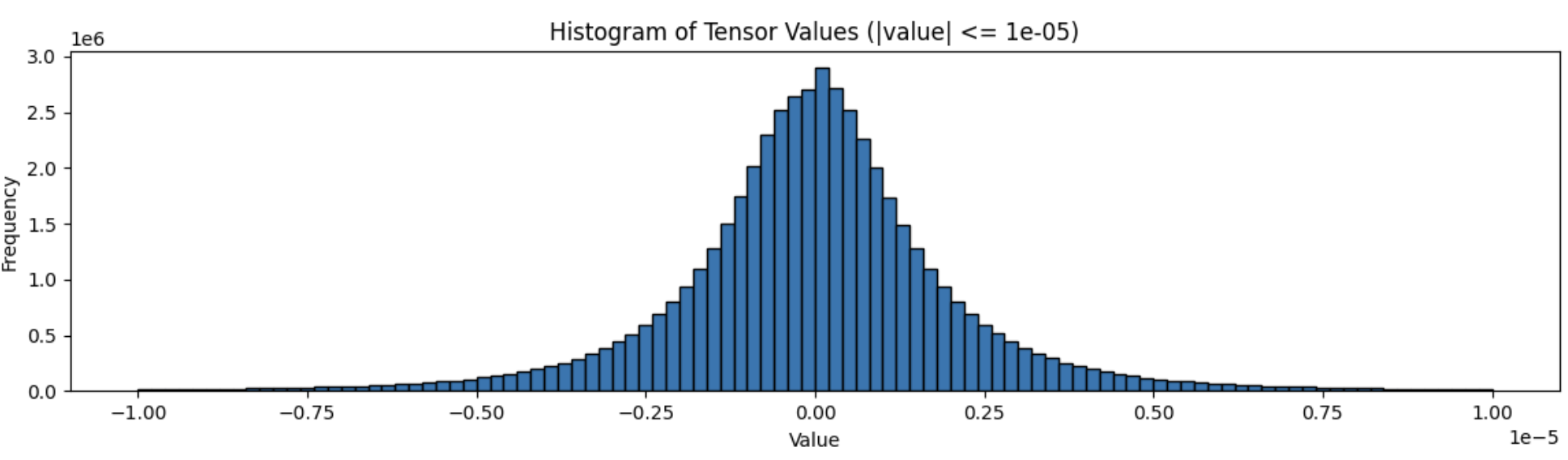}
  \caption{Histogram of tensor values}
  \label{fig:histogram_tensor_values}
\end{figure*}

Fig. \ref{fig:histogram_tensor_values} shows the histogram of optimizer tensor values, which tell us that numerical distribution of optimizer states is non-uniform, so we have to take non-linear quantization method to quantize optimizer states. Fig. \ref{fig:histogram_tensor_values} indicates that the distribution of optimizer tensor values approximately contributes to a normal distribution. This inspires us that we can split the tensor value range into many clusters utilizing the properties of normal distribution. Then elements in each cluster are balanced, now we can quantize the data in each cluster respectively.

We firstly build $m$ clusters $\textbf{C}$ in the tensor value range. Then we calculate the mean and standard deviation of the tensor values, and make the number of clusters contributes to normal distribution, which means the closer the value range nears to zero, the more the number of clusters in this range. All the tensor values will be assigned to a specified cluster and we also record the cluster labels $\textbf{L}$ that each value belongs to.

For each cluster, we use the method proposed by Dettmers \cite{dettmers_8-bit_2022} to quantize the data. The quantized value of each element in each cluster is formulated as Eq. \ref{eq:qmap-distance}

\begin{equation}
\label{eq:qmap-distance}
    T_i^Q = \arg \underset{j=0}{\overset{2^n}{min}} |\textbf{Q}_j^{map} - \frac{\textbf{T}_i-b}{S}|
\end{equation}
$\textbf{T}$ is the input tensor of each cluster, $S$ is scale factor, $b$ is offset, $Q^{map}$ is to map an n-bit integer into a real element in the domain $D$ of the target quantization data type. 

\begin{figure}[h]
  \centering
  \includegraphics[width=0.5\textwidth,height=0.4\textwidth]{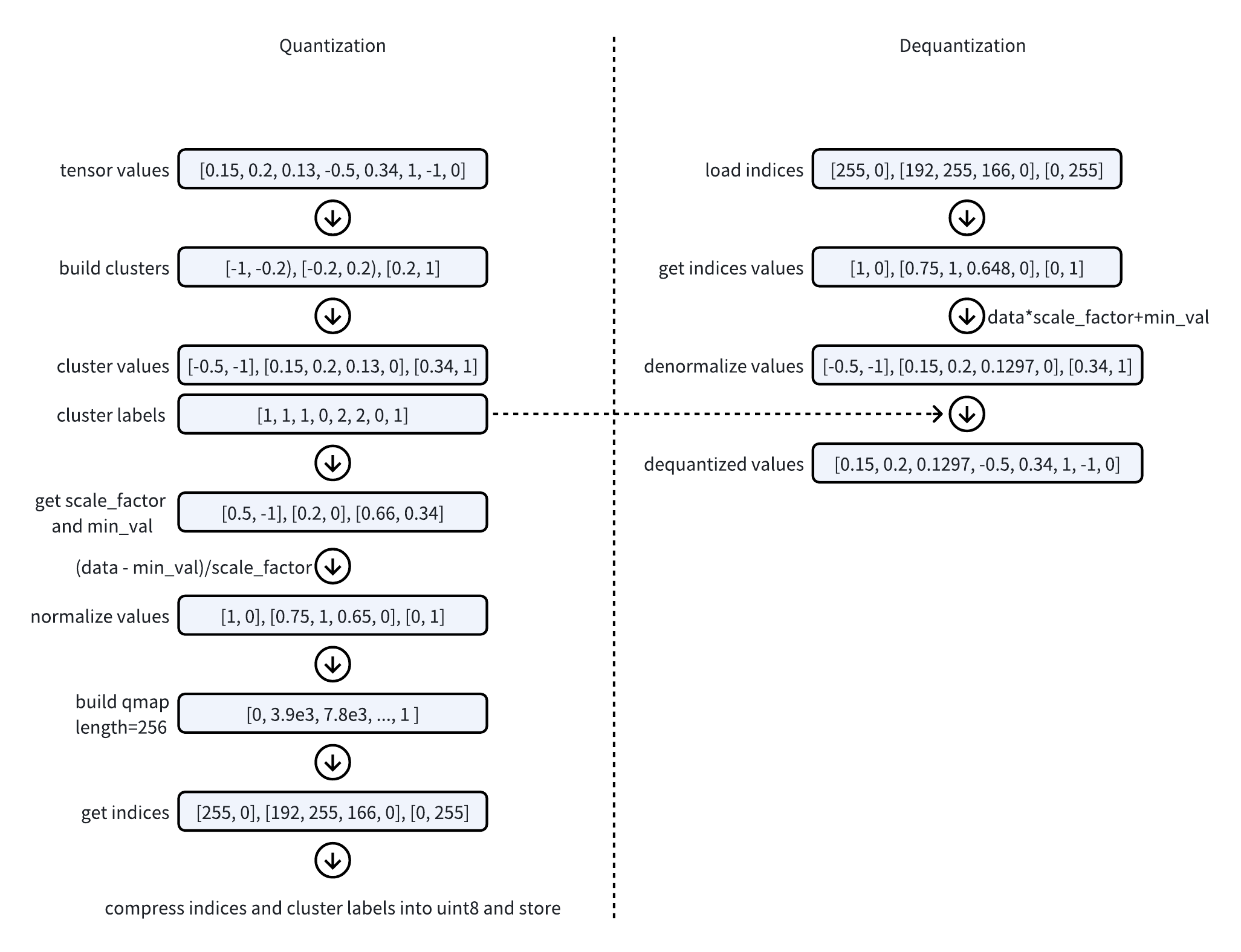}
  \caption{Cluster Based Quantization Compression}
  \label{fig:quantile-compression-method}
\end{figure}

We calculate the maximum value ($max\_val$) and minimum value ($max\_val$) in each cluster and set $S = max\_val - min\_val$, $b=min\_val$ to use asymmetric quantization. We set $k=8$ to use uint 8 as the target quantization data type. To dequantize data from $\textbf{T}_i^Q$, just calculate by $\textbf{T}_i^D = \textbf{Q}^{map}(\textbf{T}_i^Q)*S+b$. $\textbf{T}_i^D$ will be mapped back to original value $\textbf{V}_j$ by Eq. \ref{eq:decluster} where $\textbf{T}_{\textbf{C}_i}$ represents element values in cluster $i$. You can see the  quantization and dequantization procedure of optimizer states in Fig. \ref{fig:quantile-compression-method}.

\begin{equation}
\label{eq:decluster}
\begin{split}
    & \textbf{V}_j = \textbf{T}_{\textbf{C}_ik} \\
    s.t.  & \textbf{L}_{j} == i \\
    & i \in [0, m ) \\
    &  k \in [0, |\textbf{C}^i|)
\end{split}
\end{equation}




To analyze the theoretical compression ratio, we can see what we saved, that is $S$ and $b$ of each cluster, cluster labels $L$, original shape and quantized data $T^Q$. Suppose the number of elements in a tensor in $n$, and the number of clusters is $m$. The original bytes required is $4n$ bytes. In our quantization method,  $S$ and $b$ of each cluster occupy $4*2m$ bytes, $L$ occupies $\frac{log^m}{8}*n$ bytes, original shape saves the row and column of the tensor which requires $4*2$ bytes, quantized data occupies $n$ bytes. The total bytes $B$ required is $8m + (\frac{log^m}{8}+1)*n + 8$. In practice, we have tried to set $m$ to be less equal than 16 to save $L$ in uint4 data type and it proves to be effective. Now $B=\frac{3}{2}n+136$. In conclusion, our proposed cluster based quantization method will have approximately 2.67x compression ratio.

\subsection{Evaluation Metrics}

Model compression involves a trade-off among various metrics, including compression ratio, compression/decompression speed, and impact on model precision. To comprehensively evaluate the effectiveness of checkpoint compression techniques, we consider three key metrics:

\begin{itemize}
\item \textbf{Compression Ratio}: The ratio of the original model size to the compressed model size. 
\item \textbf{Compression/Decompression Speed}: The additional time required by the compression algorithm compared to the naive (no compression) method. 
\item \textbf{Impact on Model Precision}: The precision loss measured by the Mean Squared Error (MSE) between the original and decompressed models. 
\end{itemize}

We integrate these factors into a unified quality metric $Q$, defined as

\begin{equation} 
    Q = w_1 \cdot CR + w_2 \cdot CS + w_3 \cdot PS 
\end{equation}

where 

\begin{itemize}
    \item \( w_1, w_2, w_3 \) are weights reflecting the importance of each factor, typically chosen such that \( w_1 + w_2 + w_3 = 1 \). (normalized)
    \item \( CR \) is the normalized compression ratio score.
    \item \( CS \) is the normalized compression speed score.
    \item \( PS \) is the normalized precision score.
\end{itemize}

The variables \(w_1\), \(w_2\), and \(w_3\) possess distinct values under various conditions. For instance, in the training of an LLM, \(w_2\) is approximately equal to \(w_3\) and both are greater than \(w_1\). In the checkpointing process of an LLM, \(w_3\) is approximately equal to \(w_1\) and both are greater than \(w_2\).

\section{Implementation}

\subsection{Asynchronous Checkpoint Engine}

Inspired by DLRover, we build a checkpoint engine to asynchronously save model checkpoints using client-server structure. Both the client and server are written in Python. Built upon our checkpoint engine, in-memory redundancy is able to save multiple checkpoints in shared memory.

\subsection{Bitmask Based Sparsification}





To address the limitations of naive sparse matrix storage, we propose a bitmask-based sparse storage method. This technique is particularly effective when parameter changes are below 93.75\%, outperforming existing methods across a wide range of change rates.






The implementation of our bitmask-based sparse storage method involves several key steps, as described in Algo. \ref{algo:bitmask}

\begin{algorithm}
\caption{Bitmask Based Sparsification}
\label{algo:bitmask}
\begin{algorithmic}[1]
\STATE Recursively compute the differences between checkpoints
\STATE Generate a bitmask representing non-zero elements in the difference
\STATE Flatten and pad the bitmask 
\STATE Pack every 8 bits into an \textit{uint8} value 
\STATE Store the compressed bitmask and non-zero values
\end{algorithmic}
\end{algorithm}

\subsection{Cluster Based Quantization}

Our cluster based quantization method works for optimizer states, including the replication of model weights, Adam first moment estimate and Adam second moment estimate. The algorithm is described in Algo. \ref{algo:cluster-quantile}.

\begin{algorithm}
\caption{Cluster Based Quantization}
\label{algo:cluster-quantile}
\begin{algorithmic}[1]
\STATE Calculate tensor values mean $\mu$ and standard deviation $\sigma$
\STATE Build m clusters contributes to normal distribution
\STATE Assign elements to the specific cluster.
\STATE Calculate the scale\_factor and min\_val of each cluster.
\STATE Normalize elements in each cluster.
\STATE Get closest uint8 value for each element in each cluster.
\STATE Store the quantized data.
\end{algorithmic}
\end{algorithm}

\subsection{Details of Modifications of The Checkpoint Engine}

In this section, we shift our focus from the general mathematical analysis to the specific details of implementing the proposed compression strategies within the Megatron-LM framework.

To integrate these compression strategies, we modified the original checkpoint engine of Megatron-LM \cite{shoeybi_megatron-lm_2020}. Specifically, we process the \texttt{state\_dict} before storing the model and optimizer states. We introduced an environment variable named \texttt{MAX\_CACHED\_ITERATION}, allowing users to configure the interval at which checkpoints are delta-encoded.

To facilitate this, we made the following changes:
\begin{itemize}
    \item \textbf{Tracker File:} We modified the tracker file to record two additional parameters: the latest base checkpoint and the iteration number corresponding to that base checkpoint.
    \item \textbf{Checkpoint Type Indicator:} We added a new file named \texttt{type.txt} inside each checkpoint folder. This file indicates whether the checkpoint is a delta checkpoint or a full base checkpoint.
\end{itemize}

When loading a checkpoint, these parameters are used together to determine and restore the most recent checkpoint accurately.

\section{Evaluation}

\subsection{Settings}
We evaluate the effectiveness of our proposed checkpoint compression method through comprehensive experiments focusing on three key aspects: \textit{compression ratio}, \textit{time efficiency}, and \textit{impact on training convergence}. Our experiments involve models of varying scales, including GPT-2 Medium and GPT models with 0.5 billion (0.5B), 3B and 7B parameters, demonstrating the scalability and generality of our approach. We use two GPU servers, each equipped with 8 A100-80GB NVLink GPUs. We compare our method with naive 8-bit quantization. The naive 8-bit quantization just packs tensor values into range [0, 255]. To evaluate the performance of our asynchronous checkpoint engine, we record time consumed in saving checkpoint.

\subsection{Performance}
\subsubsection{Asynchronous Checkpoint Engine}

We build an asynchronous checkpoint engine upon DLRover, and the time consumed for saving checkpoint is shown in Table \ref{table:test_ckpt_time}. All the data are tested in single GPU. As computed in Megatron-LM \cite{shoeybi_megatron-lm_2020}, we can infer that the maximum number of parameters placed on A100-80GB is almost 3B. We compare BitSnap with the original checkpoint method (torch.save) used by Megatron-LM, and it shows a great speedup. By the way, we haven't tested multiple GPUs checkpointing with different parallelism (e.g. model parallel, data parallel, pipeline parallel) setup because asynchronous checkpoint is not our main contribution.

\begin{table}[h]
\centering
\caption{Time takes to save a specific GPT model in seconds}
\label{table:test_ckpt_time}
\begin{tabular}{lcccc}
\toprule
Checkpoint Method & 345M & 0.5B & 1B & 3B \\
\midrule
Megatron-LM    & 4.28  & 7.10  & 15.70  & 47.52           \\
BitSnap        & 0.58  & 0.85  & 1.35   & 4.05             \\
Speedup        & 7.38  & 8.35  & 11.63  & 11.73  \\
\bottomrule
\end{tabular}
\end{table}

\subsubsection{Compression Ratios Across Training Stages}

Our method aims to significantly reduce the storage requirements of model checkpoints without compromising model accuracy. We evaluate the compression ratios achieved under varying percentages of parameter changes between consecutive checkpoints, reflecting different training stages.

\begin{itemize}
\item \textbf{High Change Rates (up to 93.75\%)}: Even when a substantial portion of parameters change, our method maintains an effective compression ratio, reducing checkpoint sizes without loss of accuracy.
\item \textbf{Low Change Rates (3.125\%)}: In later training stages where parameter changes are minimal, our method outperforms uint16 sparse storage techniques which use COO (coordinate format) sparse matrix, achieving superior compression ratios.
\end{itemize}

\begin{figure}[ht]
    \begin{minipage}[b]{0.45\linewidth}
        \centering
        \includegraphics[width=\textwidth]{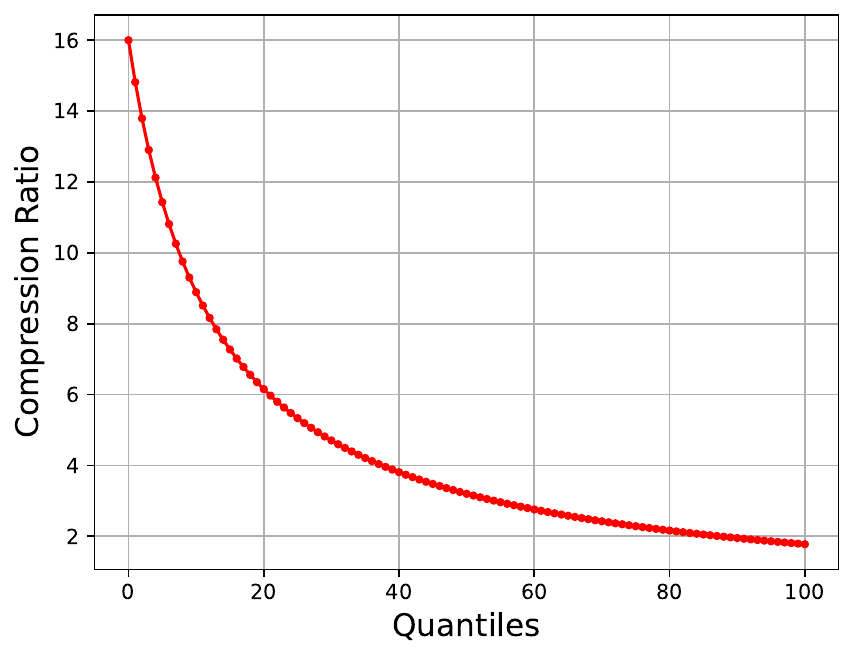}
        \caption{Compression ratio as a function of parameters changed}
        \label{fig:compression_ratio_vs_parameter_change}
    \end{minipage}
    \hspace{0.5cm}
    \begin{minipage}[b]{0.45\linewidth}
        \centering
        \includegraphics[width=\textwidth]{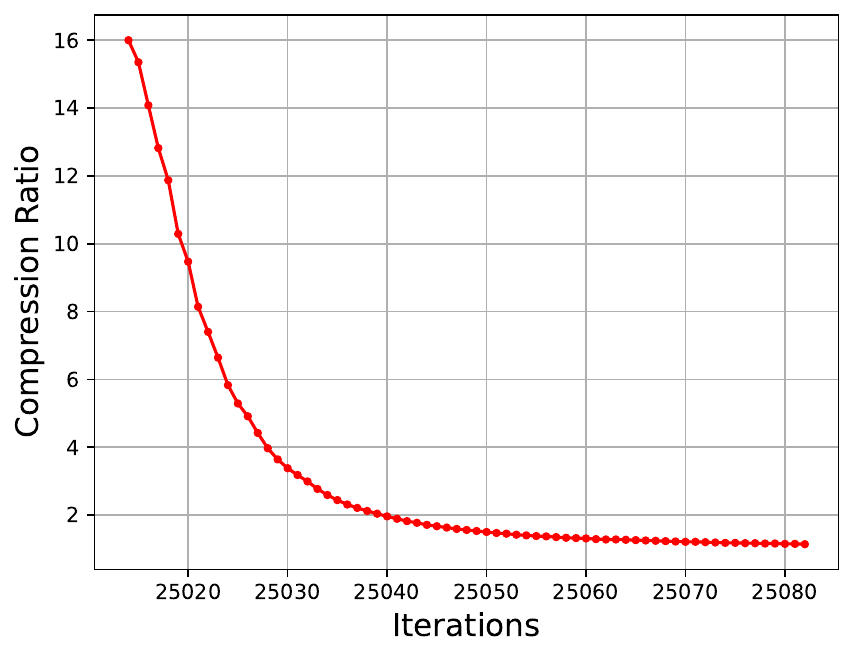}
        \caption{Compression ratio as a function of iterations changed}
        \label{fig:compression_ratio_vs_iter}
    \end{minipage}
\end{figure}

We measure compression ratio at different training stages, characterized by the proportion of parameters that change between checkpoints in Fig. \ref{fig:compression_ratio_vs_parameter_change}. And we measure compression ratio change between iterations from a base checkpoint in Fig. \ref{fig:compression_ratio_vs_iter}. The model is GPT2 Medium, and the base checkpoint is at iteration 25000. Excluding warmup iterations, the result shows we can achieve 8+x compression ratio in the next 10 iterations. The best case exists in iteration 25001 which achieves nearly the theoretical maximum compression ratio.

\subsection{Time Efficiency}

We analyze the time consumption of the key components of our method—quantization ($T_q$), clustering ($T_c$), and delta encoding—to assess time efficiency by involving multiple traning parallelism methods because it give us a better view in both compression time and traning parallelism.




\subsubsection{Impact of Parallelism}

\begin{figure}[h]
\centering
\includegraphics[width=0.5\textwidth]{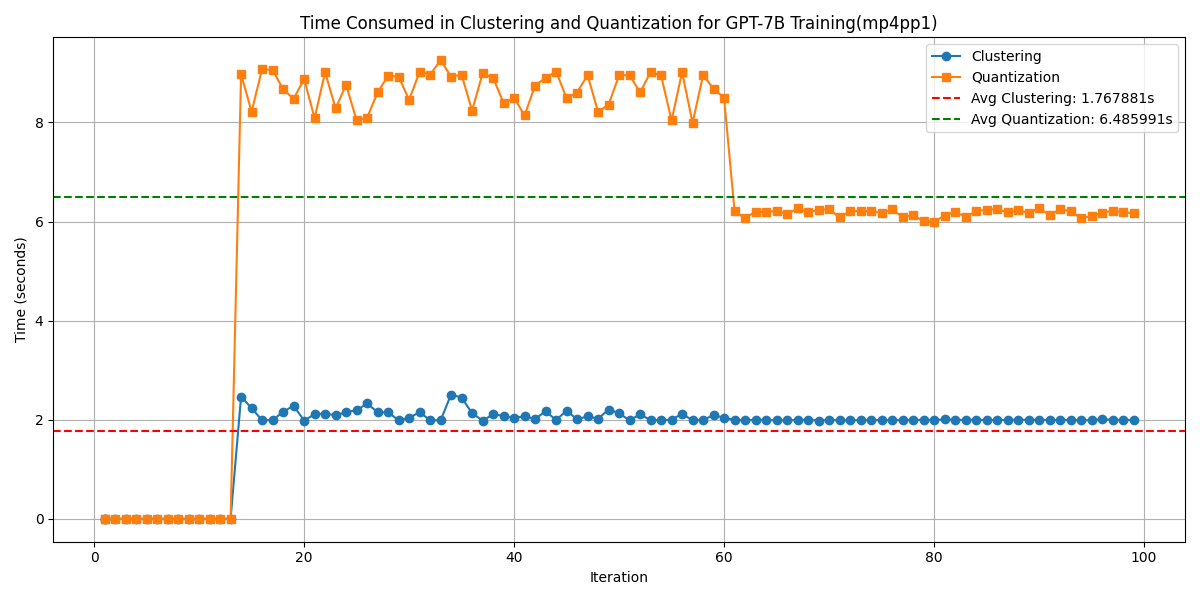}
\caption{Processing times under \textit{mp4 pp1} parallelism for the 7B model.}
\label{fig:mp4pp1_time}
\end{figure}

\begin{figure}[h]
\centering
\includegraphics[width=0.5\textwidth]{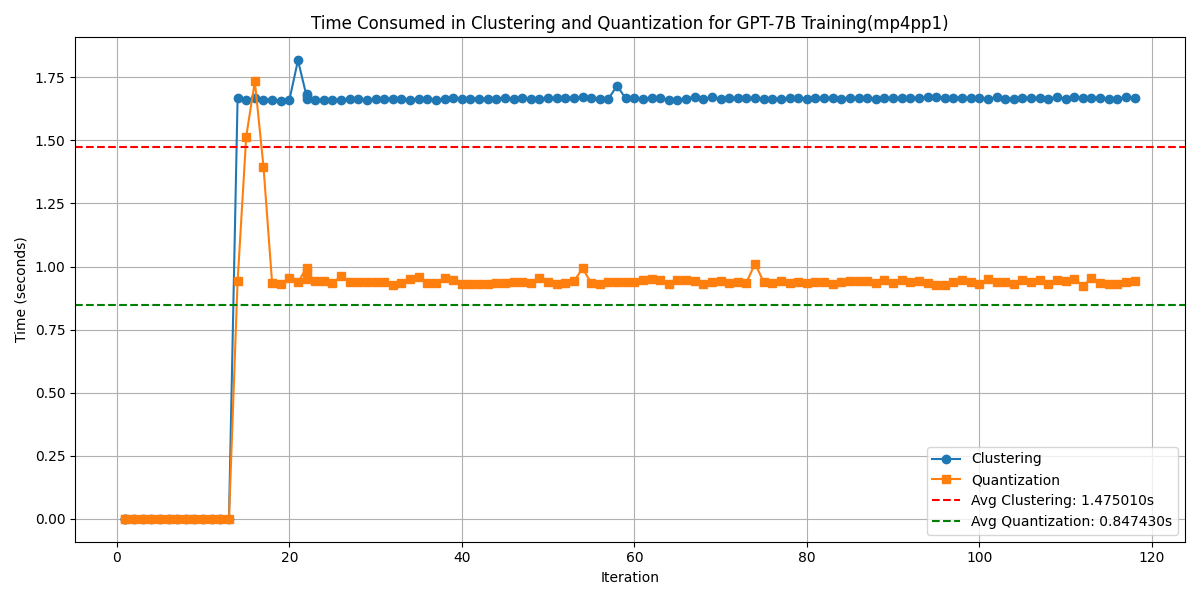}
\caption{Processing times under \textit{mp2 pp2} parallelism for the 7B model.}
\label{fig:mp2pp2_time}
\end{figure}

By integrating our method with Megatron-LM’s checkpoint engine, we leverage model and pipeline parallelism to improve efficiency. Figures~\ref{fig:mp4pp1_time} and \ref{fig:mp2pp2_time} illustrate the effect of different parallelism configurations on processing times for the 7B GPT model:

\begin{itemize}
\item \textbf{Quantization Time Reduction}: Quantization time decreases with increased parallelism due to distributed workloads.
\item \textbf{Clustering and Encoding Performance}: Both clustering and delta encoding times benefit from higher degrees of parallelism.
\end{itemize}

Pipeline parallelism further reduces processing time, indicating that our method effectively leverages parallelism to enhance scalability in large-scale training scenarios.

\subsubsection{Delta Encoding Time During Retraining}

We observe an increase in delta encoding time during retraining from a breakpoint, attributed to warm-up phases. However, delta encoding is generally not performed during retraining from a breakpoint, so this issue can be avoided. Optimizing this aspect remains an area for future improvement.

\subsection{Impact on Training Convergence}

A crucial criterion for our compression method is that it should not adversely affect model training convergence or accuracy.

\begin{figure}[h]
\centering
\includegraphics[width=0.5\textwidth]{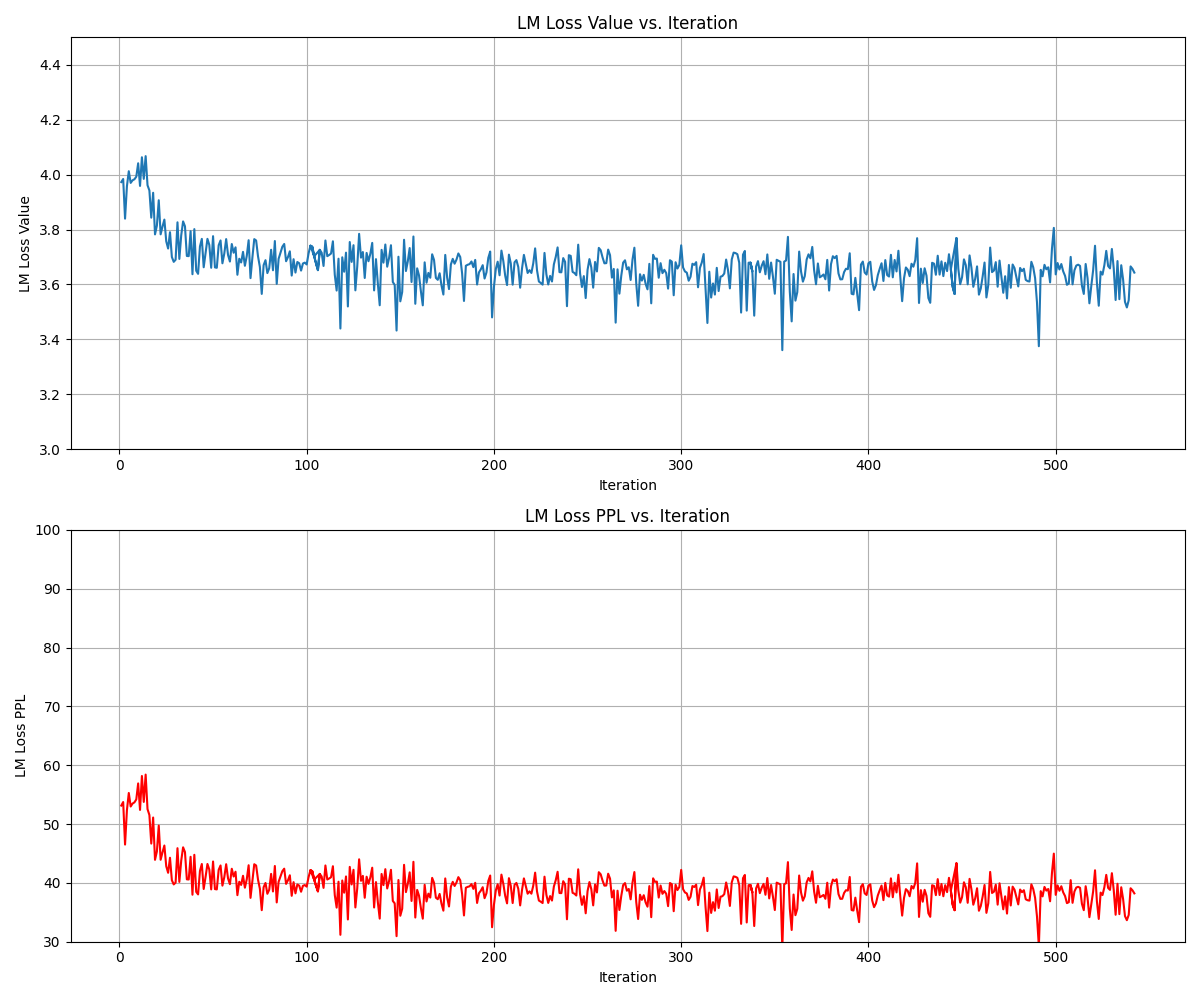}
\caption{LM loss over training iterations, comparing sparsified and unsparsified checkpoints.}
\label{fig:loss_curve}
\end{figure}

\textbf{Loss Curve for Bitmask Based Sparsification.} We track the GPT2 Medium loss over 500 iterations during the later stages of training to assess the impact of our method. As shown is Fig. \ref{fig:loss_curve}, our method maintains a smooth loss curve, indicating stable training. Unlike aggressive pruning methods that cause abnormal loss increases~\cite{li_excp_2024}, our compression does not introduce training instability. The model’s performance remains consistent with the uncompressed baseline, confirming that our compression is lossless with respect to model accuracy.

\begin{figure}[h]
\centering
\includegraphics[width=0.5\textwidth]{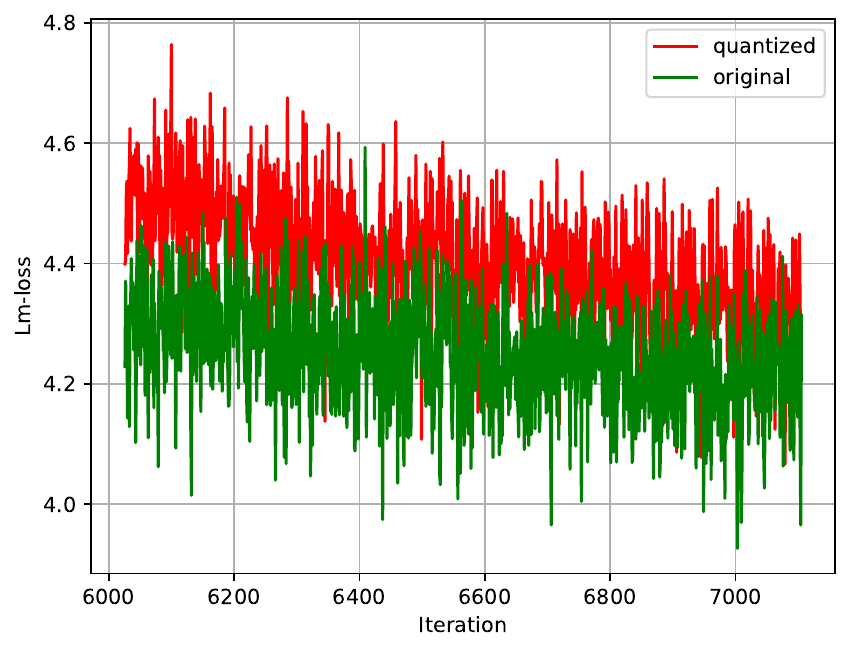}
\caption{LM loss over training iterations, comparing quantized and inquantized checkpoints}
\label{fig:loss_curve_quantized}
\end{figure}

\textbf{Loss Curve for Cluster Based Quantization.} We evaluate the GPT2 Medium loss curve when retraining from a quantized checkpoint. We save the checkpoint at iteration 6000, and load the checkpoint to recover training. We can see in Fig. \ref{fig:loss_curve_quantized} that after loading from quantized data, it has approximately 4.5\% impact on loss convergence. We calculate the Mean Relative Error (MRE) and Mean Squared Error (MSE) of the dequantized optimizer states over the original value, see Table. \ref{table:mre_mse}, it proves our quantized method has little impact on model accuracy.

Additionally, we compare our method with naive 8-bit quantization on GPT2 Medium in Table. \ref{table:mre_mse_8bit}. Our method is much more robust than the naive 8-bit quantization.

\begin{table}[h]
\centering
\caption{MRE and MSE of dequantized optimizer states over LLM models. Adam1 means Adam first moment estimate and Adam2 means Adam second moment estimate}
\label{table:mre_mse}
\begin{tabular}{lcccc}
\toprule
Metrics & 345M & 0.5B & 1B & 3B \\
\midrule
Adam1-MRE    & 9.86  & 9.89  & 10.30  & 10.32          \\
Adam1-MSE    & 1.57e-9  & 1.68e-9  & 1.78e-9   & 1.82e-9      \\
Adam2-MRE    & 0.18  & 0.20  & 0.24  & 0.25          \\
Adam2-MSE    & 1.51e-14  & 1.53e-14  & 1.56e-14   & 1.60e-14      \\
\bottomrule
\end{tabular}
\end{table}

\begin{table}[h]
\centering
\caption{MRE and MSE compared with naive 8-bit quantization on GPT2 Medium}
\label{table:mre_mse_8bit}
\begin{tabular}{lcccc}
\toprule
Metrics & BitSnap & Naive 8-bit \\
\midrule
Adam1-MRE    & 9.86  & 401188.01          \\
Adam1-MSE    & 1.57e-9  & 3.90e-8   \\
Adam2-MRE    & 0.18  & 0.11         \\
Adam2-MSE    & 1.51e-14  & 6.43e-13     \\
\bottomrule
\end{tabular}
\end{table}

Our evaluation demonstrates that the proposed bitmask-based sparsification method effectively reduces checkpoint sizes and accelerates the checkpointing process without impacting model accuracy or training convergence. And our cluster-based quantization method quantized the 32-bit optimizer states into 8-bit, and it proves to be efficient.

\section{Conclusion}

We introduced BitSnap, a system for asynchrounous checkpoint saving and loading with checkpoint compression. Our proposed bitmask based sparsification and cluster based quantization methods show a great improvement in checkpoint compression with 16x and 2x compression ratio respectively, and have little impact on precision loss. Based on our checkpoint compression methods, our in-memory redundancy method makes little pressure on memory consumption.
\nocite{langley00}

\bibliography{main}
\bibliographystyle{mlsys2025}

\end{document}